\documentclass[pdflatex,sn-mathphys-num]{sn-jnl}

\usepackage{graphicx}%
\usepackage{multirow}%
\usepackage{amsmath,amssymb,amsfonts}%
\usepackage{amsthm}%
\usepackage{mathrsfs}%
\usepackage[title]{appendix}%
\usepackage{xcolor}%
\usepackage{textcomp}%
\usepackage{manyfoot}%
\usepackage{booktabs}%
\usepackage{algorithm}%
\usepackage{algorithmicx}%
\usepackage{algpseudocode}%
\usepackage{listings}%

\usepackage{tikz}
\usepackage{pgfplots}

\usepackage{booktabs}

\begin{document}

\title{Optimal Transport-based Permutation-Invariant Bayesian Optimization of Offshore Wind Farm Layouts}


\author[1]{\fnm{Antonio} \sur{Candelieri}}\email{antonio.candelieri@unimib.it}

\author[2,3]{\fnm{Laurens} \sur{Bliek}}\email{l.bliek@tue.nl}

\affil[1]{\orgdiv{
Department of Economics Management and Statistics}, \orgname{University of Milano-Bicocca}, %\orgaddress{\street{Street}, 
\city{Milan}, 
%\postcode{100190}, \state{State}, 
\country{Italy}}

\affil[2]{\orgdiv{Department of Industrial Engineering and Innovation Sciences}, \orgname{Eindhoven University of Technology}, %\orgaddress{\street{Street}, 
\city{Eindhoven}, 
%\postcode{100190}, \state{State}, 
\country{the Netherlands}}

\affil[3]{
%\orgdiv{Department of Industrial Engineering and Innovation Sciences},
\orgname{Eindhoven AI Systems Institute}, %\orgaddress{\street{Street}, 
\city{Eindhoven}, 
%\postcode{100190}, \state{State}, 
\country{the Netherlands}}

\abstract{
Bayesian Optimization (BO) is widely and successfully adopted for solving optimization problems having an expensive-to-evaluate, black-box, and non-convex objective function. However, the \textit{vanilla} BO algorithm is not able to exploit possible symmetries  characterizing the target problem. An intuitive case is given by \textit{optimal location} problems, whose decision variables refer to a finite set of points within a continuous space, with the order of points not affecting the value of the objective function. We refer to this setting as \textit{optimization over layouts} to distinguish from \textit{optimization over point-clouds} where, instead, the order of points counts. As an instance of optimization over layouts we consider a real-life industrial-relevant application, that is the optimization of the layout of an offshore wind farm: given identical wind turbines, switching any pair of them has not any effect on the annual energy production. Based on Optimal Transport theory, we propose a Permutation-Invariant BO approach, namely PIBO, proved to provide better wind farm layouts when compared to the vanilla BO approach while cutting computation time roughly in half.}


\maketitle

\section{Introduction}

\subsection{Problem statement and motivations}
This paper addresses global optimization of a black-box, expensive-to-evaluate, non-convex function over \textit{layouts}. With \textit{layout} we refer to a point-cloud whose points permutation does not affect the value of the objective function. A more general definition could be \textit{permutation-invariant} optimization, which means that the objective function does not depend on any permutations of its inputs or subsets of them. Without loss of generality, we anyway prefer to introduce the term \textit{layout} because it is more consistent with the specific application considered in this paper -- as well as many other real-life applications -- but also use permutation-invariant as a synonym.

From a practical perspective, many real-life applications that involve \textit{optimal placement in a continuous space} fall into the proposed formulation, such as sensor placement for environmental monitoring~\cite{Hellan2023BayesianOA}, well placement for carbon capture and storage~\cite{Fotias2024OptimizationOW}, integrated circuit design~\cite{Deshwal2021BayesianOO}, and optimization of wind farm layouts~\cite{EXPObench}.

%, and routing problems~\cite{Xie2025FromSA} --> not continuous

Denote by $P\in \mathbb{R}^{m\times d}$ a \textit{point-cloud} consisting of $m$ points (e.g., a set of $m$ sensors, wells, pumps, etc.) within a $d$-dimensional space and by $f:\Omega\subset\mathbb{R}^{m\times d}\rightarrow \mathbb{R}$ a \textit{scoring} function to maximize. Optimizing \textit{over point-clouds} means solving:
\begin{equation}
    \label{eq:point-clouds1}
    \underset{P\in\Omega\subset\mathbb{R}^{m\times d}}{\max}\; f(P)
\end{equation}
A naive way to proceed is to consider a \textit{vectorization} $\varphi:\Omega\rightarrow\Omega_v\subset\mathbb{R}^{md}$ of the point-clouds, and solve the following transformed problem:
\begin{equation}
    \label{eq:point-clouds2}
    \underset{v_P\in\Omega_v\subset\mathbb{R}^{md}}{\max}\; f(\varphi^{-1}(v_P))
\end{equation}
with $\varphi^{-1}(v_P)$ denoting the inverse of vectorization.

This assumes that the order of the points in the point-clouds -- that is, the order of the rows in $P$ and the order of the coordinates in $v_P$ -- counts.
However, if the order does not count (as in the applications mentioned above), then $\forall P \in \Omega, f(P)=f(\varsigma(P))$, with $\varsigma(P)$ denoting any permutation of the rows of $P$. This is dramatic because even if the target problem has only one global optimal layout, formulations (1) and (2) have $m!$ (i.e., $m$-factorial) global optima.

This undesired ``replication mechanism'' does not affect only the global optimum, but every possible solution, leading $f$ to be ``infested'' by local (and global) optima due to the replication of the true underlying function, which is instead defined over layouts.

Since $f$ is assumed to be black-box, expensive-to-evaluate, and non-convex, Bayesian Optimization (BO)~\cite{archetti2019bayesian,garnett2023bayesian} would be the best choice to address the problem, but the undesired replication mechanism can definitely prevent \textit{vanilla} BO from quickly converging towards an optimal solution. Moreover, if the feasible region is a sub-region of the typically box-bounded search space $\Omega$, which is the case for \emph{constrained BO}, we see that $f$ becomes even more complicated, with replications also affecting the feasible region.

\begin{figure}[h!]
    \centering
    \includegraphics[width=0.45\linewidth]{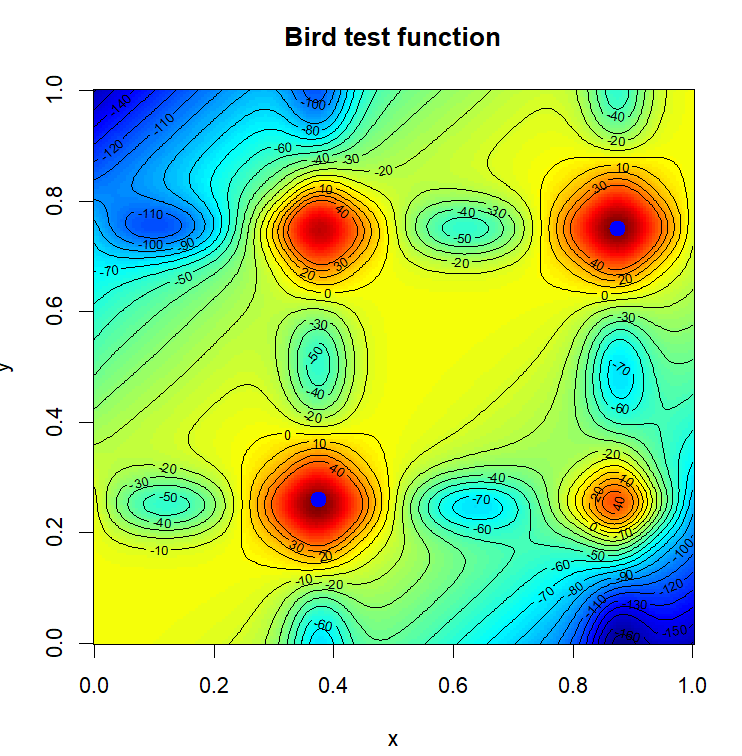}
    \includegraphics[width=0.45\linewidth]{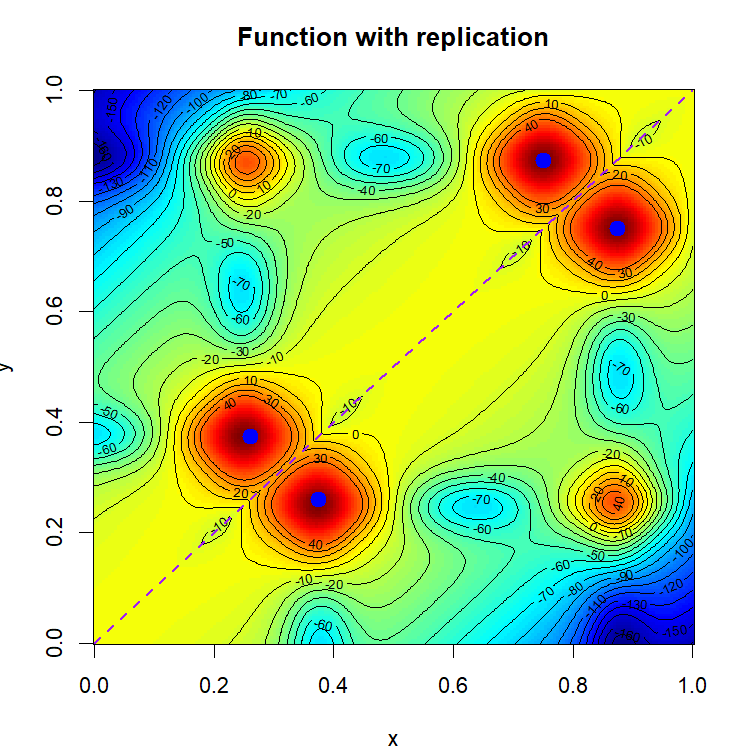}
    \caption{A simple example of the undesired replication: on the left, the Bird test function (with search space rescaled in $[0,1]^2$ and changed in sign for considering maximization); on the right, a function over layouts (aka permutation-invariant function) obtained by mirroring the original Bird function from below to above the bisector (i.e., $f(x_1,x_2)=f(x_2,x_1)$). Although the true function \textit{over layouts} is just the one below the bisector, with just two global optima, optimizing over point-clouds, like in formulation \ref{eq:point-clouds1} and \ref{eq:point-clouds2}, leads to consider a function with four global optima (i.e., replication mechanism).}
    \label{fig:1}
\end{figure}

For simplicity, here we introduce an example with $m=2$ and $d=1$. Indeed, the aim is to search for a couple of points in a 1-dimensional space that minimize a function $f$ such that $f(x_1,x_2)=f(x_2,x_1)$. In this example, we obtain $f$ from the well-known Bird test function. Figure \ref{fig:1} shows: on the left the original 2-dimensional Bird function (changed in sign to consider maximization instead of minimization and with search space scaled to $[0,1]^2$), on the right our function $f$. It is evident that the function above the bisector is a replica of the one below, and, consequently, the two global optima of the original Bird function are also replicated, leading to four global optima for our function $f$. Obviously, one could also consider to replicate the function above the bisector, but in this case we lose the global optima of the original Bird function, leading to a useless example.

Although one can think that more global optima means more chances to find one of them, this is not true because the replication mechanism could lead to an extremely ``wiggling'' $f$, with global optima becoming ``needles-in-the-haystack'', a well-known condition that prevents convergence of BO \cite{bull2011convergence,wang2014theoretical,wabersich2016advancing,berkenkamp2019no,candelieri2024mle}. Surely, the proposed example is quite trivial, but it should help to understand that what happens to a $(m\times d)$-dimensional function is even harder to imagine.

\subsection{Main contribution}

We make use of the Optimal Transport (OT) theory to achieve a permutation-invariant version of BO, which we call PIBO. The basic idea is to refer every point-cloud to a reference one -- suitably sampled outside the search space -- and to compute the optimal transport from the reference to the specific point-cloud representing a candidate solution. Under certain assumptions, OT theory ensures that every reference point is associated to one point of the candidate solution uniquely -- or almost always uniquely -- and that this association can be expressed in terms of a \textit{flow} that is nothing else than a vector. Indeed, BO will be applied over this \textit{space of flows} and where any flow, given the reference point-cloud, can be transformed into layout -- instead of a point-cloud -- ensuring permutation invariance.

\noindent
The main contributions of this paper can be summarized as follows:
\begin{itemize}
    \item We propose a permutation-invariant BO approach, namely PIBO, based on OT theory for addressing optimization problems over point-clouds in a continuous space, but where the order of points in the cloud does not account. Indeed, we refer to optimization over layouts;
    \item We evaluate PIBO on a real-life relevant application, specifically the optimization of an offshore wind farm's layout;
    \item We empirically prove that PIBO outperforms other not-permutation-invariant BO algorithms, in terms of optimal value of the objective function as well as computational efficiency.
\end{itemize}

To ensure replicability and encourage the adoption of PIBO, our code is freely available on~\url{https://github.com/lbliek/PIBO-for-Wind-Farm-Layout-Optimization}.

\subsection{Related works}

Permutation invariance in BO has been considered before, typically in the context of BO for combinatorial optimization, since permutations are common in a large variety of combinatorial optimization problems, such as routing, planning, and scheduling.
In~\cite{Deshwal2021BayesianOO}, the standard kernels used in Gaussian Processes (GPs) for BO were adapted to two different kernels: the Kendall kernel and the Mallows kernel, both of which take into account permutations. For the Kendall kernel, the acquisition function optimization problem was modeled into a quadratic assignment problem, which was solved using semi-definite programming. For the Mallows kernel, the acquisition function was optimized with a local search heuristic that takes into account permutations.
In~\cite{Irurozki2021UnbalancedMM}, similar kernels were used in the unbalanced Mallows model, combined with the Borda ranking algorithm.
The same model was applied later in the asteroid routing problem~\cite{LopezIbanez2022TheAR}.
\citet{Xie2025FromSA} extended these kernels to the high-dimensional setting, reducing the computational complexity of the kernels from $O(n^2)$ to $O(n \log n)$ using the Merge kernel.
Optimal transport theory has also been used in BO for neural architecture search~\cite{Kandasamy2018NeuralAS}, where the layout of a neural network was optimized using a distance metric similar to the earth mover's distance or the Wasserstein distance.

The previous works all consider purely combinatorial optimization problems such as the traveling salesperson problem or the quadratic assignment problem, or other routing or scheduling problems.
In contrast, several works exist that consider a permutation-invariant case where the relation between points is continuous, however, in these works the number of possible locations is still finite or discrete.
In~\cite{Garnett2010BayesianOF}, a sensor set selection problem was solved, where sensors can be placed according to a finite set of possible locations. Even though the combinatorial search space consists of over 2 million possibilities, this remains different from a purely continuous problem.  Their solution is based on the earth mover's distance, which is related to the Wasserstein or Mallows distance, and they solve the related linear program with the Hungarian algorithm.
A similar sensor placement problem for environmental monitoring was presented in~\cite{Hellan2023BayesianOA}, and the authors provide a problem generator based on real air pollution data, with a finite set of possible locations in 2D space.
Perhaps more similar to our work is  \citet{Fotias2024OptimizationOW} on well placements for carbon capture and storage. While the relation between points is continuous just like in the sensor placement problems, this work discretizes the search space into cells. The authors calculate the distance to the origin for all cells, which is unique for every cell in their problem, and use this to get to a linear program in the form of an assignment problem that they solve with the Hungarian algorithm. They also consider the situation where there are two different types of objects to place, complicating the problem further.

Moreover, permutation invariance in continuous space is also a challenging research topic in Deep Learning (DL) \cite{elaarabi2025adaptive,perez2025quantification}, especially for Graph Neural Networks (GNNs) \cite{balan2203permutation,liu2025rotation}, Transformers \cite{patil2025permutation,selvaraj2025permutation}, and more recently Physics-Informed Networks \cite{chen2026permutation}. In all these studies, the learning process must ensure that the output of the neural network is invariant to permutations of the input neurons, leading to a permutation-invariant optimization of the training loss function.  

To the best of our knowledge, the only work that considers permutation-invariant BO where both the relation between points as well as the space of possible points is continuous, is BO with approximate set kernels~\cite{Kim2021BayesianOW}. In this work, the set kernel is approximated using a random scalar projection. Ordering the resulting scalars is a trivial task that drastically reduces computational complexity. The algorithm is shown to outperform a vectorized approach on synthetic problems, as well as on the problem of selecting initial solutions for a clustering algorithm.

\section{Methods}

\subsection{Bayesian Optimization in brief}
Bayesian Optimization (BO) is a sample-efficient sequential model-based method for global optimization of black-box, expensive-to-evaluate, non-convex objective functions \cite{garnett2023bayesian,archetti2019bayesian}.
In a generic iteration, BO performs two actions: (\textit{a}) fitting a probabilistic surrogate model of the objective function, depending on the current set of observations $\mathcal{D}=\left\{(x^{(i)},y^{(i)})\right\}_{i=1:n}$, and (\textit{b}) selecting the next \textit{query} by optimizing an acquisition function that relies on the probabilistic surrogate model and balances between exploitation (i.e., local search) and exploration (i.e., global search). Then, the next query $x'$ is evaluated and $\mathcal{D}$ is updated accordingly, as $\mathcal{D}\leftarrow\mathcal{D} \cup \left\{(x',y'=f(x'))\right\}$.

The Gaussian Process (GP) \cite{gramacy2020surrogates,williams2006gaussian} is the most widely adopted option for the probabilistic surrogate model, providing both a prediction for every new input $x$ and the associated predictive uncertainty, denoted by $\mu(x|\mathcal{D})$ and $\sigma(x|\mathcal{D})$, respectively. These two quantities also depend on a \textit{kernel} (aka \textit{covariance function)} chosen a-priori and whose hyperpaparemeters are typically tuned via Maximum Likelihood Estimation (MLE) or Maximum-A-Posteriori (MAP), which can be considered as a penalized-MLE. Different kernels induce different GP models, with Exponential and Squared Exponential kernel representing the two most different possibilities: continuous but not every-where differentiable GP's prediction (the first) and continuous and infinitely differentiable (the second).

In terms of acquisition function, there exists a wide range of possibilities, usually organized into two families: \textit{improvement-based} and \textit{entropy-based} \cite{shahriari2015taking}. Those in the first family are designed to search for $f(x^*)$, while others are designed to search for $x^*$. Although apparently irrelevant, this distinction leads to completely different strategies, with information-based acquisition functions usually offering a greater sample efficiency, but entailing a computational burden so large for the determination of the next query to make them useful only when the cost of evaluating $f(x)$ is significantly higher. In this paper, we consider one of the most widely used acquisition functions: the GP's Upper Confidence Bound (GP-UCB), namely $\mu(x|\mathcal{D}) + \xi \sigma(x|\mathcal{D})$, with $\xi$ managing the so-called \textit{uncertainty bonus}. GP-UCB is an \textit{optimistic-in-face-of-uncertainty} acquisition function, promoting more risk-seeking queries with increasing $\xi$.

\subsection{Optimal Transport theory in brief}

Optimal Transport (OT) theory \cite{peyre2019computational,OTambrogio} is a mathematically principled framework to compare two probability measures by \textit{transporting}, at minimum cost, the \textit{source} probability measure in order to match the \textit{target} one. Formally, it means solving
\begin{equation}    \mathcal{L}_c(\mu,\nu)=\underset{\pi\in\Pi(\alpha,\beta)}{\min} \int_{U\times V} c(u,v)\text{ d}\pi(u,v)
    \label{eq:OT}
\end{equation}

where $\alpha\in\mathcal{P}_2(U)$ and $\beta\in\mathcal{P}_2(V)$ are two probability measures, with finite second moments, respectively defined in $U\subseteq\mathbb{R}^d$ and $V\subseteq\mathbb{R}^d$, and where $c(u,v)$ is the cost of transporting a unit of probability mass from $u\in U$ to $v\in V$, and $\pi$ belongs to the set of all possible joint probability measures having $\alpha$ and $\beta$ as marginals, as denoted by $\Pi(\alpha,\beta)$.
Importantly, when $c(x,y)$ is a \textit{distance metric}, $\mathcal{L}_c(\alpha,\beta)$ is also a distance metric. Specifically, if $c(x,y)=\|x-y\|^p_p$, the equation (\ref{eq:OT}) becomes the so-called $p$-Wasserstein distance, with $p\in[1,\infty]$:
\begin{equation}
    \mathcal{W}_p(\alpha,\beta) = \underset{\pi\in\Pi(\alpha,\beta)}{\min}\ \left[\int_{U\times V}\|u-v\|_p^p \text{ d}\pi(u,v)\right]^{\frac{1}{p}}
    \label{eq:Wp}
\end{equation}
Another important OT's property is the possibility to compare two continuous, two discrete, as well as one continuous and one discrete, probability distributions.

In this paper, we consider the 2-Wasserstein distance (i.e., $p=2$) between two empirical discrete distributions, also called point-clouds, meaning that $U=\left\{u^{(i)}\right\}_{i=1:n}$ and $V=\left\{v^{(j)}\right\}_{j=1:m}$, and $\alpha=\frac{1}{n}\sum_{i=1:n}\delta_{u^{(i)}}$ and $\beta=\frac{1}{m}\sum_{j=1:m}\delta_{v^{(j)}}$, with $\delta_z$ denoting the Dirac's delta function centered at location $z\in\mathbb{R}^d$.\\

According to the Brenier theorem (Remark 2.24 in \cite{peyre2019computational}), when one of the two probability measures to compare is absolutely continuous and has finite second moment, then (\ref{eq:Wp}) admits a unique optimal \textit{transportation plan} $\pi^*$ and there exists a unique optimal \textit{transportation map} $T^*$ that solves the following problem
\begin{equation}
    \underset{T_\#\alpha=\beta}{\min}\int_{\mathbb{R}^d}\left\|u-T(u)\right\|^2\text{d}\alpha(u)
\end{equation}

where $T:\mathbb{R}^d\rightarrow\mathbb{R}^d$, $T_\#\alpha=\beta$ means that the total probability mass of $\alpha$ must be transported to match $\beta$, and $\alpha$ is assumed to the absolutely continuous probability measure.
Moreover, the optimal transportation plan $\pi^*$ and the optimal transportation map $T^*$ are related to each other by the relationship
\begin{equation}
    \pi^*=(\text{Id},T^*)_\#\alpha
    \label{eq:equivalence}
\end{equation}
with ``Id'' denoting the identity function.

Unfortunately, the situation changes if both probability measures are not absolutely continuous -- like in our case. Then, there still exists an optimal transport plan $\pi^*$, but it is not unique, and the existence of an optimal transport map $T^*$ is not guaranteed. However, if there exists a map $T$ such that $\beta=T_\#\alpha$ and the optimal plan $\pi^*$ is $\pi^*=(\text{Id},T)_\#\alpha$, then $T$ is an optimal map. The opposite does not hold: if $T$ is an optimal map, then the plan $\pi=(\text{Id},T)_\#\alpha$ fulfills the constraint related to the marginal, but must not be optimal.\\

Luckily, our setting has some useful properties that allow us for simplification. First, since we are considering point-clouds with $m$ elements each, we can consider the \textbf{Monge's formulation} of the OT problem between two empirical discrete distributions, that is:
\begin{equation}
    \varsigma_* \in \underset{\varsigma \in S_m}{\min} \sum_{i=1:m} \|u_i-v_{\varsigma(i)}\|^2
\end{equation}
where $\varsigma \in S_m$ denotes one among the $m!$ permutations of the $m$ rows of the target point-cloud $V$. This is an assignment problem, which is a combinatorial optimization problem. Since the transportation cost is a convex function (i.e., the squared Euclidean distance) the existence of the optimal solution is guaranteed, but not its uniqueness.

A practical method for solving Monge's problem is considering its linear relaxation, also known as \textbf{Kantorovich's formulation}:
\begin{align}
    T_* \in \underset{T \in \mathbb{R}_{\geq 0}^{m\times d}}{\arg \min} & \quad \sum_{i,j=1:m} C_{ij}^2 T_{ij}\\
    s.t. &  \quad T\mathbf{1}_m = \frac{1}{m}\mathbf{1}_m\\
    & \quad T^\top\mathbf{1}_m = \frac{1}{m} \mathbf{1}_m\\
    & \quad T_{ij}\geq 0,\;\forall\;i,j=1:m
\end{align}
where $\mathbf{1}_m$ denotes an $m$-dimensional all-ones column vector, the entries of the cost matrix $C\in\mathbb{R}^{m\times m}$ are $C_{ij}=\|u_i-v_j\|$.

If $T_{*ij}\in \{0,1\}\; \forall\; i,j=1:m$, then $T_*$ is a permutation matrix and $T_*^\top U=\varsigma_*(V)$, where $\varsigma_*$ is an optimal solution of the Monge's problem.

More importantly, in the Kantorovich formulation we can also establish whether $T_*$ is unique or not: it is unique when the matrix $C$ is in \textit{the general position}. An important empirical result is that when the transport cost is convex and the point-clouds are not on a prefixed grid -- that is exactly our setting -- $C$ is \textit{almost surely} in general position for every source and target point-clouds. In any case, we have to choose our reference point-cloud $\bar{P}$, thus we can always do that to have every $C$ matrix in a general position.
Another practical consideration is that it is not necessary to ensure uniqueness for all possible solutions: even if there are a number of replicas, it will be significantly less than $m!$.

Finally, if $T_*$ is such that $T_*^\top U = \varsigma_*(V)$, then we can express $\varsigma_*(V)$ as $U+X_*$ and, consequently, $X_*=\varsigma_*(V)-U=T_*^\top U-U=(T_*^\top - I)U$. We can call $X_*$ the \textit{optimal flow} from $U$ to $V$, and this is a crucial concept in our approach because, to ensure permutation-invariance, we are going to perform \textit{BO over the space of optimal flows} instead of, directly, the space of point-clouds.

\subsection{BO over layouts: permutation-invariance through OT theory}

In this section, we summarize the main modifications we propose to the vanilla BO framework to deal with permutation invariance via OT theory.

First, a reference point-cloud $\bar{P}$ is sampled from a multivariate normal distribution (i.e., bivariate in our real-life application) with mean and covariance matrix such that the points in $\bar{P}$ are outside the problem's domain (i.e., the physical space relate to the deployment area in our real-life application). This is important to ensure that any cost matrix between $\bar{P}$ and a $P \in \Omega$ is almost surely in general position.

In every iteration $n$, a set of optimal flows $\mathbb{X}=\left\{X_*^{(h)}\right\}_{h=1:n}$ is available, along with the function values of the associated layouts stored in the set $\mathbf{y}=\left\{y^{(h)}=f(\bar{P}+X_*^{(h)})\right\}_{h=1:n}$, where $\bar{P}+X_*^{(h)}$ is the $h$-th layout.

The peculiarity of our approach is that the GP model is fitted on the optimal flows $\mathbb{X}$ instead of the associated layouts, so that the acquisition function is be optimized over this space.
We denote the posterior mean of the GP by $\hat{f}(X|\mathbb{X},\mathbf{y})\simeq f(\bar{P}+X)$ and the associated predictive uncertainty by $\sigma(X|\mathbb{X},\mathbf{y})$. As an acquisition function, we choose the GP-UCB, namely $\bar{f(X|\mathbb{X},\mathbf{y})}+\xi \sigma(X|\mathbb{X},\mathbf{y})$, with different values for the uncertainty-bonus $\xi$.
Later in the text, we explain how the acquisition function is optimized, since we must include operational constraints related to the real-life application addressed in this study. At this stage, it is more important to clarify that as the new $X'$ is provided by GP-UCB, the associated $P'=\bar{P}+X'$ is computed. Since $X'$ could not be an optimal flow between $\bar{P}$ and $P'$, a Kantorovich's problem between these two point-clouds is solved to obtain $T_*'$ (and it is also checked that $T_{*ij}'\in\{0,1\},\; \forall i,j=1:m$). Therefore, $X_*'=(T_*'^\top-I)\bar{P}$ is added to the set $\mathbb{X}$ and $y'=f(T_*'^\top\bar{P})$ to $\mathbf{y}$. This ensures us that the GP model and the outcome of the acquisition always lie in a subregion of the entire search space (i.e., optimal flow space) with no replications\footnote{Looking at the simple example in Figure \ref{fig:1}, this means that every solution will lay in the region below the bisector,}.

Like in the basic BO algorithm, the process iterates until a maximum number of layouts has been evaluated.

\section{Application to wind farm layout optimization}

As a real-life test case, we consider a problem in the offshore wind energy domain.
The annual energy production (AEP) of an offshore wind farm is highly dependent on the location of the wind turbines. A wrong wind farm layout can reduce the AEP due to wake effects under extreme wind conditions. However, calculating these wake effects requires expensive physics simulators, making the identification of an optimal layout of the wind turbine farm an expensive optimization problem.

In~\cite{EXPObench}, the optimization of the layout of a wind farm was addressed through a variety of surrogate-based optimization algorithms with default configurations for $m=5$ wind turbines and $5$ random yearly wind scenarios. In that work, permutation invariance was not considered.

To speed up our experiments, instead of a physics simulator, we use an ensemble of predictive Machine Learning models trained on data from~\cite{EXPObench}, as detailed in Section~\ref{sec:surr}. This ensemble takes approximately $50$ ms to evaluate one layout, on our hardware, while physics simulators can easily require hours or days.

Given a layout of the wind farm as input, the ensemble provides a reliable estimate of the AEP in GWh/year, which must be maximized. The area where the wind turbines are to be installed -- also called \textit{physical space} -- is box-bounded and preliminarily rescaled to $[0,1]^{m\times 2}$, with $m$ the number of identical wind turbines. Overall, this leads to the following maximization problem:
\begin{align}
    \underset{P \in [0,1]^{m\times 2}}{\max}& \quad f(P)\\
    \text{s.t.} & \quad \|p_i-p_j\|\geq \rho,\; \forall\;i,j=1:m
\end{align}
where $p_i$ and $p_j$ refer to two rows of $P$ (i.e., they are the coordinates of two wind turbines) and $\rho$ is twice the diameter of the turbine's rotor. 
The constraint is included for practical safety reasons: any two wind turbines need a minimum distance $\rho$ between them, which is exactly twice their rotor diameter. The original simulator would fail if this constraint is violated, while the objective in~\cite{EXPObench} would simply return $0$. On the other hand, our ensemble of Machine Learning models could predict a value different from $0$ -- due to prediction error. Since the closed-form expression of the constraint is available and does not entail any simulation/prediction, we directly include and evaluate it in the optimization of the acquisition function in BO.

As it is assumed that all wind turbines are identical, swapping any pair of them has no effect on the objective function, neither constraint, other than any noise that is present in the computation of the objective function itself. As such, this is a permutation-invariant optimization problem, where we must search over optimal layouts instead of point-clouds. 

Moreover, since in our approach every point-cloud $P$ is expressed as an optimal flow from the reference one, namely $P=\bar{P}+X$, the constraints can be rewritten as:
\begin{equation}
    \|\bar{p}_i+x_i-\bar{p}_j-x_j\| \geq \rho 
    \label{eq:14}
\end{equation}

Although in \textbf{Algorithm \ref{alg:1}}, which describes our approach, we use this formulation of the constraints, the following (not-immediately-intuitive) Proposition is used in the implementation of the approach.\\

\noindent
\textbf{Proposition 1.} \textit{If $\|\bar{p}_i-\bar{p}_j\|<\rho\;\forall i,j=1:m$ then the constraint (\ref{eq:14}) can be simply rewritten as $\|x_i-x_j\|\geq\|\bar{p}_i-\bar{p}_j\|+\rho$.}\\

\noindent
\textit{Proof.} First, rewrite the constraint (\ref{eq:14}) as $\|\bar{p}_i+x_i-\bar{p}_j-x_j\|^2\geq \rho^2$, then we have:
\begin{equation}
    \|\bar{p}_i-\bar{p}_j\|^2 + \|x_i-x_j\|^2 - 2\|\bar{p}_i-\bar{p}_j\|\|x_i-x_j\|\geq \rho^2
\end{equation}
Now, replace $\|x_i-x_j\|$ with $z$ and solve the inequality:
\begin{equation}
    z^2 - 2\|\bar{p}_i-\bar{p}_j\| z + \|\bar{p}_i-\bar{p}_j\|^2-\rho^2 \geq 0
\end{equation}
then it follows that
\begin{equation}
    \begin{cases}
        z \leq \|\bar{p}_i - \bar{p}_j\| - \rho\\
        z \geq \|\bar{p}_i + \bar{p}_j\| + \rho\\
    \end{cases}
\end{equation}\\
\noindent
However, since $z\geq 0$ by definition (i.e., $z=\|x_i-x_j\|$), choosing $\|\bar{p}_i-\bar{p}_j\| < \rho$ implies $z\geq \|\bar{p}_i-\bar{p}_j\|+\rho$, meaning:
\begin{equation}
    \|x_i-x_j\| \geq \|\bar{p}_i-\bar{p}_j\| + \rho
\end{equation}
\hfill QED\; $\blacksquare$

\begin{algorithm}[h!]
\caption{PIBO: Permutation Invariant Bayesian Optimization}\label{alg:1}
\begin{algorithmic}
\\
\State \textbf{Input:} $n_0 \geq 2m+1, N>n0$\\

\State $\mathbb{P}=\left\{P^{(h)}\in [0,1]^{m\times d}\right\}_{h=1:n_0}$ and $ \mathbf{y}=\left\{y^{(h)}=f(P^{(h)})\right\}_{h=1:n_0}$
\State $\bar{P}\sim\mathcal{N}(\mathbf{m}_P,\Sigma_P): \forall h=1:n_0,\; T_*^{(h)}$ is a permutation matrix and $C^{(h)}$ is in general position.
\State $\mathbb{X}=\left\{X_*^{(h)} = (T_*^{(h)\top}-I)\bar{P}\right\}_{h=1:n_0}$
\State $n\gets n_0$\\

\While{$n < N$}\\
   \State fit a GP with $\mu(X|\mathbb{X},\mathbf{y}) \simeq f(\bar{P}+X)$ and $\sigma(X|\mathbb{X},\mathbf{y})$ the predictive uncertainty\\
   \State $X' \in \underset{X\in \mathbb{R}^{m\times d}}{\max}\;\mu(X|\mathbb{X},\mathbf{y}) + \xi\; \sigma(X|\mathbb{X},\mathbf{y})$
   \State \quad \quad \quad s.t.\quad $\|\bar{p}_i+x_i-\bar{p}_j-x_j\|\geq \rho$\\

   \State compute $T_*'$ between $\bar{P}$ and $\bar{P}+X'$\\

   \State $\mathbb{X} \gets \mathbb{X} \cup \left\{(T_*'^\top - I)\bar{P}\right\}$
   \State $\mathbf{y} \gets \mathbf{y} \cup \left\{f(T_*^\top \bar{P})\right\}$
   \State $n\gets n+1$\\
   
\EndWhile\\

\State \textbf{Output:} $(X^+,y^+) \in (\mathbb{X},\mathbf{y}): y^+ = \underset{n=1:N}{\min}\left\{y^{(n)}\right\}$
\end{algorithmic}
\end{algorithm}

\newpage

\section{Experiments and Results}

\subsection{Experimental setting}
Our setting consists of $m=5$ identical wind turbines that must be optimally placed in a box-bounded area, preliminary rescaled to $[0,1]^2$. The minimum distance between two wind turbines, in the rescaled space, is $\rho=0.1512$.

Our PIBO algorithm is compared against other baseline approaches: all evaluate 500 solutions, i.e. point-clouds of $m=5$ points each. PIBO is the only permutation-invariant approach. The entire set of approaches is:
\begin{itemize}
    \item \textbf{Latin Hypercube Sampling (LHS) within the physical space:} 5 LHS points within $[0,1]^{m\times 2}$, repeated 100 times;
    \item \textbf{Uniform random sampling (URS) within the physical space:} 5 points within $[0,1]^{m\times 2}$, repeated 100 times;
    \item \textbf{Tree-Parzen Estimator (TPE) working over the physical space:} initialized with $2m+1=11$ point-clouds, uniformly sampled at random; 500 point-clouds overall; 
    \item \textbf{LHS within the space of flows:} 5 LHS points within the space of flows, repeated 100 times; 
    \item \textbf{vanilla BO over the space of flows:} initialized with $2m+1=11$ optimal flows computed from as many point-clouds randomly initialized via LHS in the physical space; 500 flows overall;
    \item \textbf{vanilla BO over the physical space:} initialized with $2m+1=11$ point-clouds generated via LHS in the physical space; 500 point-clouds overall.  
\end{itemize}
To mitigate the impact of randomness, all the mentioned configurations have been run 30 times starting from as many different seeds.

The three BO approaches share the same initialization in each independent run. Two different GP kernels are considered, namely Exponential and Squared Exponential, and four different values for the uncertainty bonus in GP-UCB, specifically $\xi\in\{0,1,3,6\}$. The acquisition function is optimized, inexactly, \cite{kim2025bayesian} through 10'000 LHS solutions in the algorithm-specific search space. 

TPE %~\cite{Bergstra2011AlgorithmsFH} 
is a close-to-BO approach: the main difference is that TPE does not rely on any probabilistic surrogate model: instead of modeling the objective function, TPE models the distributions of ``good'' and ``bad'' solutions. The specific TPE implementation we have used is from the \emph{hyperopt} package~\cite{Bergstra2013MakingAS} as part of the \emph{EXPObench} library~\cite{EXPObench} in python. Contrary to all the other methods, TPE cannot deal with the constraint, thus we have used a workaround: we simply return $0$ for the objective function in case of a constraint violation.

\subsection{Empirical results}

Table~\ref{tab:baselineresults} summarizes the results for the baseline methods, namely the different types of random search and TPE. Remarkably, TPE is the baseline method that provides, on average, the highest value of the objective function (and also the highest maximum value over the 30 independent runs). Thus, a sequential approach obviously provides better results than simple random search methods.

\begin{table}[h!]
\caption{Value of the optimal solution (average and standard deviation over 30 independent runs) provided by the baseline approaches (i.e., not relying on any probabilistic surrogate model): the higher the better. Percentage of feasible solutions (i.e., not violating the constraint) is also reported.}
\label{tab:baselineresults}
\begin{tabular}{lcc}
%{p{0.3\textwidth}p{0.2\textwidth}p{0.1\textwidth}p{0.12\textwidth}}
\toprule
\textbf{Baseline} & $y_{best}$ \textbf{avg (sd)} & \textbf{ \% feasible solutions}\\
\midrule
LHS in the Physical Space & 69.27 (0.63)  & 94\% \\
LHS in the Flow Space & 68.39 (1.24) & 56\% \\
URS in the Physical Space & 68.71 (1.24) & 51\% \\
TPE in the Physical Space & \textbf{69.75} (0.62) & 57\% \\
\bottomrule
\end{tabular}
\end{table}

\newpage 

Table~\ref{tab:mainresults} summarizes the results of the three BO methods (one per column) with respect to the different kernel types and $\xi$ values in GP-UCB (one configuration per row). The most relevant considerations can be summarized as follows:
\begin{itemize}
    \item all BO implementations outperform TPE (i.e., the best baseline method);
    \item \textit{vanilla BO on point-clouds} never provides a better solution than the other two BO implementations. This means that \textbf{targeting the optimization problem in the space of flows is more effective than in the physical space}.
    \item PIBO with Exponential kernel offers better solutions, on average, than \textit{vanilla BO on flows} -- and, obviously, \textit{vanilla BO on point-clouds}. Indeed, \textbf{the permutation-invariance at the core of PIBO allows it to beat its competitors}.
    \item When the Squared Exponential kernel is considered, \textit{vanilla BO on flows} behaves better than PIBO, but performances are anyway lower than those obtained for the Exponential kernel counterparts, for the same value of $\xi$.
    \item Overall, PIBO with the Exponential kernel and $\xi=6$ offers the best performance.  
\end{itemize}

\begin{table}[h!]
\centering
\resizebox{\textwidth}{!}{
\begin{tabular}{lcccc|cc}
    \toprule
    \textbf{GP's} & $\xi$ \textbf{in} & \textbf{PIBO (ours)} & \textbf{BO on flows} & \textbf{BO on point-clouds} & \textbf{runs won} & \textbf{runs won by}\\
    \textbf{kernel} & \textbf{UCB} & $y^+$ \textbf{avg (sd)} & $y^+$ \textbf{avg (sd)} & $y^+$ \textbf{avg (sd)} & \textbf{by PIBO} & \textbf{BO on flows} \\
    \midrule
    \textbf{Exp.} & 0 & 70.38 (0.32) & \textbf{70.56} (0.41) & 69.86 (0.50) & 12/30 & 17/30 \\
                  & 1 & \textbf{70.60} (0.36) & 70.46 (0.40) & 69.95 (0.40) & 16/30 & 13/30 \\
                  & 3 & \textbf{70.63} (0.38) & 70.58 (0.38) & 69.80 (0.37) & 17/30 & 13/30 \\
                  & 6 & \underline{\textbf{70.77}} (0.33) & 70.66 (0.45) & 69.82 (0.26) & 15/30 & 15/30 \\
    \midrule
    \textbf{Sq. Exp.} & 0 & 70.00 (0.39) & \textbf{70.13} (0.52) & 69.29 (0.66) & 9/30 & 19/30 \\
                      & 1 & 69.93 (0.50) & \textbf{70.03} (0.61) & 69.32 (0.70) & 12/30 & 15/30 \\
                      & 3 & 69.90 (0.53) & \textbf{69.96} (0.53) & 69.47 (0.59) & 14/30 & 13/30 \\
                      & 6 & 70.04 (0.60) & \textbf{70.13} (0.51) & 69.33 (0.62) & 14/30 & 13/30 \\
    \bottomrule
\end{tabular}
}
\caption{Value of the optimal solution (average and standard deviation over 30 independent runs) for the three BO approaches: the higher the better. Best value for each pair kernel type - $\xi$ value is in bold; the best overall is also underlined. Number of runs won by PIBO and \textit{vanilla BO on flows} are also reported (runs won by \textit{vanilla BO on point-clouds} is obtained by difference).}
\label{tab:mainresults}
\end{table}

\begin{figure}[h!]
    \centering
    \includegraphics[width=0.9\linewidth]{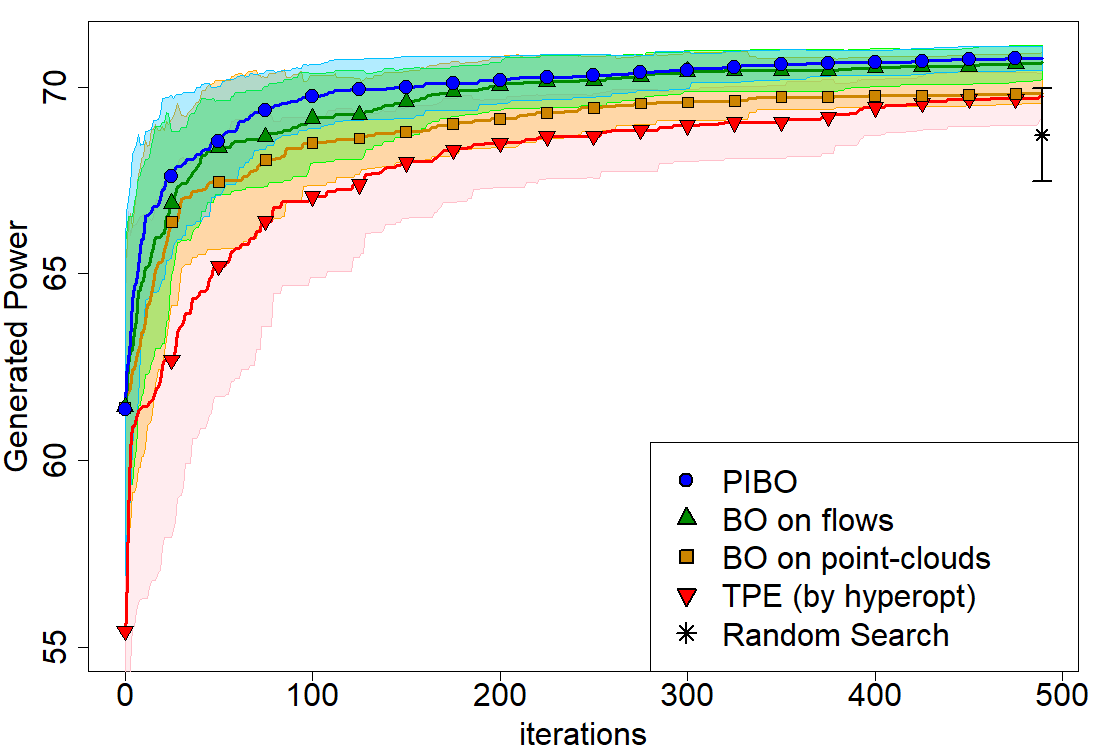}
    \caption{Comparing convergence of the three algorithms (PIBO, vanilla BO on flows, and vanilla BO on point-clouds) with Exponential kernel for the GP and $\xi=6$ in GP-UCB. Results are averaged on 30 independent runs. Important remark: for each run the three algorithms share the same set of initial solutions, randomly chosen via LHS within the physical space: \textit{vanilla BO on point-clouds} keeps going by working in the same space while PIBO and \textit{vanilla BO on flows} works on the space of flows. TPE do not share the same set of initial solutions because the hyperopt library -- which implements it -- only allows for uniformly random initialization in the physical space.}
    \label{fig:convergence_plot}
\end{figure}

We now focus on the configuration of kernel type and $\xi$ in the GP-UCB which has provided the best performance, specifically the Exponential kernel and $\xi=6$. Figure \ref{fig:convergence_plot} shows the evolution of the \textit{best seen} (i.e., the best value observed up to each iteration): PIBO outperforms all other methods on average. Only \textit{vanilla BO on flows} is able to reduce the gap to PIBO, but after 200 iterations, and in any case providing, on average, a lower best seen. On the contrary, \textit{vanilla BO on point-clouds} behaves more similarly to TPE than the other two BO approaches. It is important to clarify that the initial best seen for TPE is different from BO because TPE is an off-the-shelf algorithm and we cannot control its internal initialization. Anyway, it quickly converges to values close to those of \textit{vanilla BO on point-clouds}, leading to the conclusion that working in the physical space prevents the convergence towards better solutions.

\newpage 

Finally, the comparison between the three BO approaches can be considered as a sort of \textit{ablation study}:
\begin{itemize}
    \item when we move from the optimization over point-clouds, within the physical space, to the optimization over flows, \textit{vanilla BO on flows} outperforms \textit{vanilla BO on point-clouds}, meaning that the problem in the space of flow is -- in some sense -- ``easier'' for BO
    \item when we move further from the optimization over flows to the optimization over \textit{optimal} (in OT terms) flows, then PIBO outperforms \textit{vanilla BO on flows}, meaning that the permutation-invariance implied by OT allows to further simplify the BO process in this specific search space.
\end{itemize}

\begin{figure}[h!]
    \centering
    \includegraphics[width=0.32\linewidth]{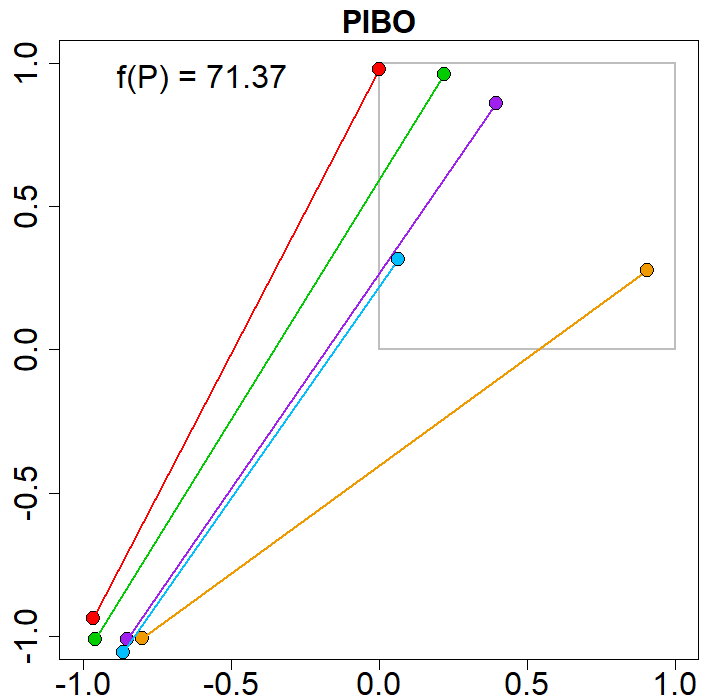}
    \includegraphics[width=0.32\linewidth]{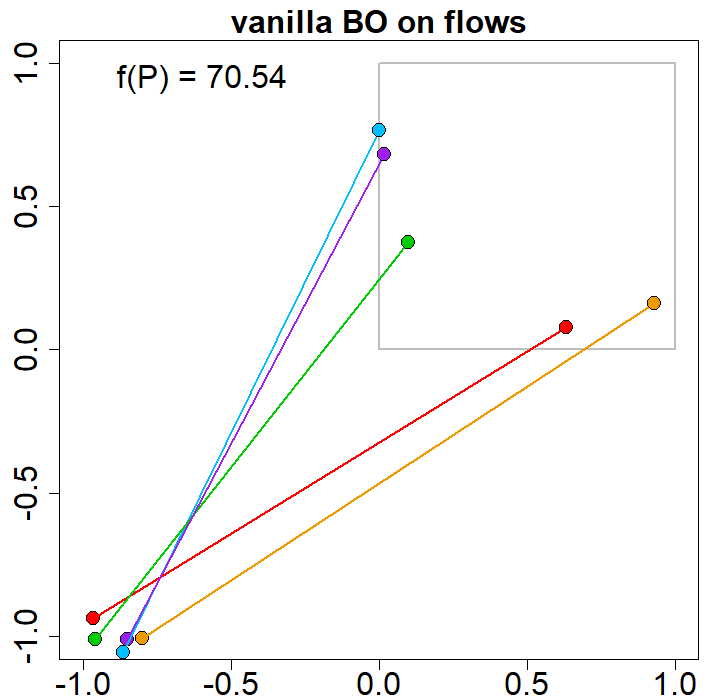}
    \includegraphics[width=0.32\linewidth]{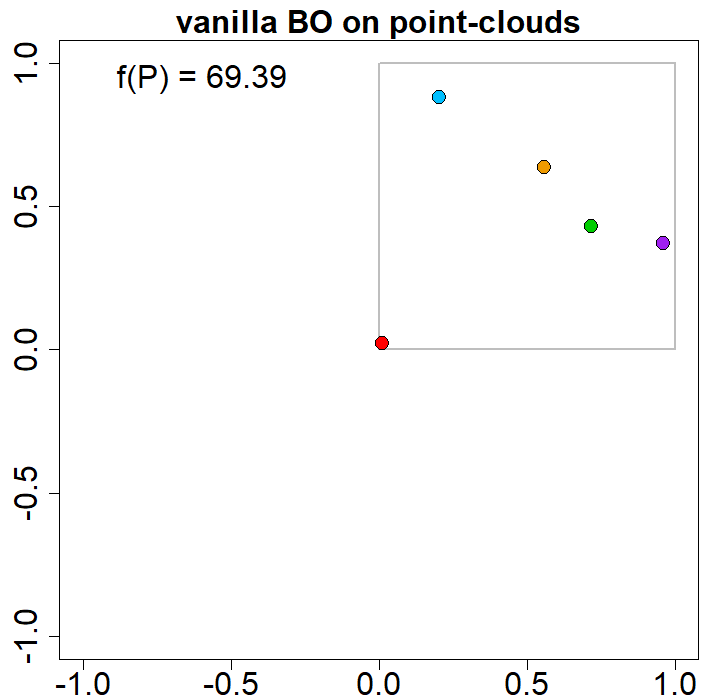}
    \caption{Best solutions identified by the three BO algorithms, for the same run. GP's kernel is Exponential and $\xi=6$ in GP-UCB. The gray square represents the physical search space (i.e., the area where wind turbines must be placed); the five points at the left-bottom corner represents the reference point cloud $\bar{P}$ (omitted in the rightmost picture because \textit{vanilla BO on point-clouds} does not involve OT theory). PIBO (on the left) provides the highest value for the objective function and the associated solution is consistent with OT (i.e., no crossing flows). \textit{Vanilla BO on flows} (in the middle) provides the second best solution among the three, but the placement is not consistent with OT (indeed, we observe crossing flows). Finally, \textit{vanilla BO on point-clouds} (on the right) provides the worst solution among the three.}
    \label{fig:single_best_solution}
\end{figure}

The crucial role of permutation-invariance is more evident in Figure \ref{fig:single_best_solution} that compares the best optimal wind farm layouts provided by PIBO, \textit{vanilla BO on flows}, and \textit{vanilla BO on point-clouds}, with respect to one of the 30 independen runs. The GP kernel is Exponential and $\xi=6$ in GP-UCB. Each wind turbine is uniquely identified by a color, the box in the upper-right part of each picture is the area where wind turbines must be placed (i.e., the physical space), the small point-cloud in the lower-left part of the pictures is the reference point-cloud $\bar{P}$ (omitted in the rightmost picture because \textit{vanilla BO on point-clouds} does not rely on flows and, thus, it does not use $\bar{P}$). Finally, the flow from each reference point to the associated turbine is depicted by a line of the same color as the turbine.

In the PIBO's solution (leftmost picture) the flows are optimal in OT terms (i.e., there are no crossing flows), contrary to those in \textit{vanilla BO on flows} (picture in the middle). This means that the \textit{vanilla BO on flows} is not permutation-invariant. Finally, \textit{vanilla BO on point-clouds} (rightmost picture) behaves even worse, with a significantly lower value of the objective function.

\begin{figure}[h!]
    \centering
    \includegraphics[width=0.32\linewidth]{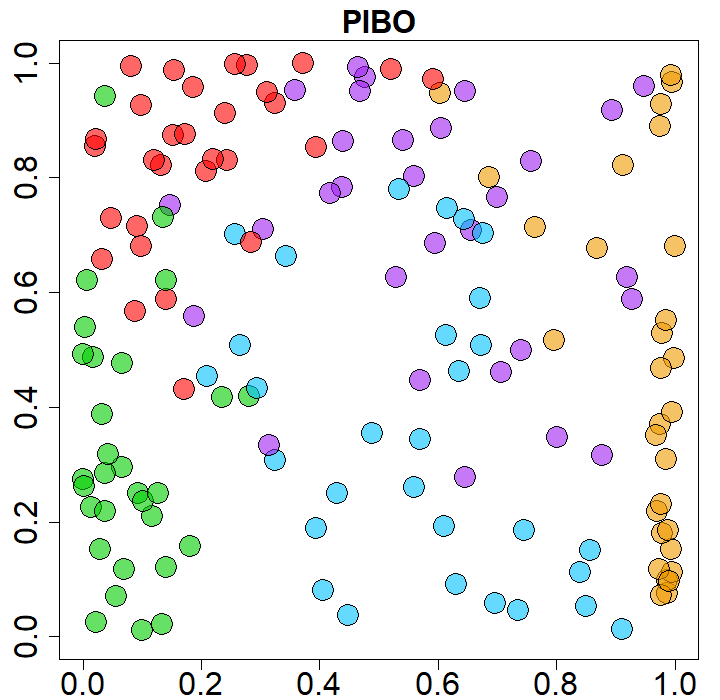}
    \includegraphics[width=0.32\linewidth]{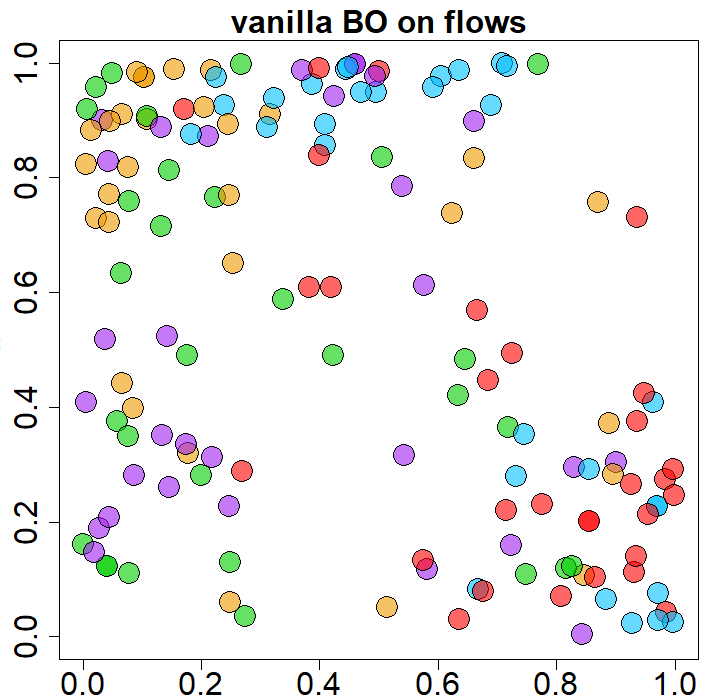}
    \includegraphics[width=0.32\linewidth]{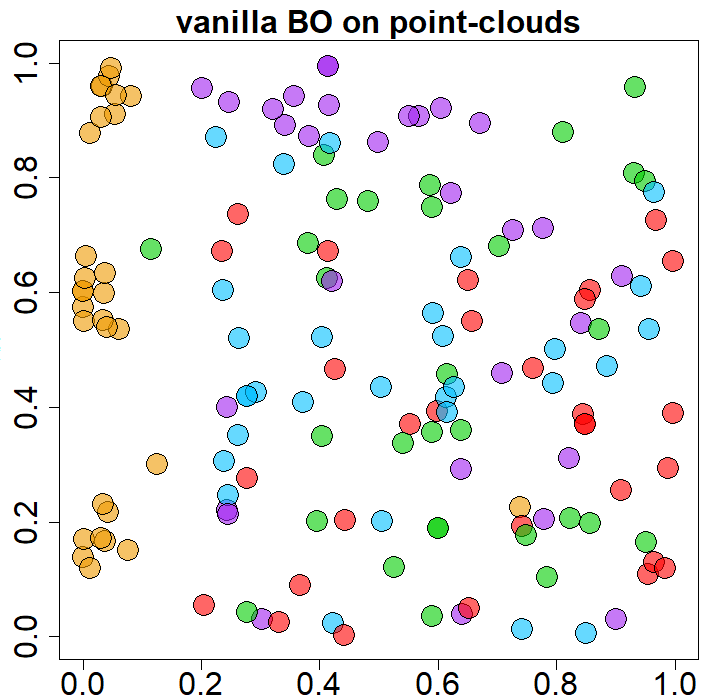}
    \caption{Optimal wind farm layouts provided by the three BO implementations, separately. Each color is associated to one of the $m=5$ wind turbines, every point refers to the optimal location of that turbine for one of the 30 independent runs. Thanks to permutation-invariance at its core, PIBO is the only approach able to suggest a non-linear organization of the Physical Space into (possibly overlapping) \textit{influence regions}, one for each wind turbine.}
    \label{fig:best_solutions}
\end{figure}

Another relevant consideration on the importance of permutation-invariance comes from the analysis of the best wind farm layouts provided by the three BO implementations on the 30 independent runs, overall (again under Exponential kernel and $\xi=6$ in GP-UCB).
In Figure \ref{fig:best_solutions}, where each wind turbine is uniquely determined by a color, it is evident that the optimal layouts provided by PIBO suggest a sort of (non-linear) organization of the physical space into \textit{influence regions}, possibly overlapping, individually associated with the $m=5$ wind turbines. No one of the two vanilla BO implementations can infer such an organization, because of their inability in dealing with permutation-invariance of the wind turbines' locations. 

\newpage

Compared with baseline methods, the three BO implementations have a computational overhead due to the GP fitting and optimization of the acquisition function. In this study, the overhead is almost completely due to the GP fitting, since the acquisition function is optimized inexactly \cite{kim2025bayesian}. Moreover, PIBO requires solving a linear programming problem (i.e., Kantorovich's problem) at every iteration, leading to it possibly being slower than the other two BO implementations. Surprisingly, this is not the case: permutation-invariance at the core of PIBO simplifies the GP fitting step, improving computational efficiency. Specifically, we have registered the following times\footnote{Experiments have been run on 13th Gen Intel(R) Core(TM) i9-13900H (2.60 GHz), 64.0GB of RAM, Microsoft Windows 11.} (GP fitting plus acquisition) for each one of the three BO implementations (average $\pm$ standard deviation on the 30 independent runs):
\begin{itemize}
    \item PIBO: 347.949 $\pm$ 86.125 [secs] 
    \item vanilla BO on flows: 692.062 $\pm$ 172.817 [secs]
    \item vanilla BO on point-clouds: 654.688 $\pm$ 116.558 [secs]
\end{itemize}

While computation time is comparable between \textit{vanilla BO on flows} and \textit{vanilla BO on point-clouds}, it is approximately half for PIBO. Since the computation time is almost completely due to GP fitting, this empirically proves our initial motivation: the replication mechanism dramatically and drastically affects the GP fitting because we are trying to model an extremely wiggling function infested by global and local optima. On the contrary, dealing with permutation-invariance -- as PIBO does -- allows one to model the true and smoother underlying function defined over layouts instead of point-clouds.

\section{Conclusions}
We have proved that using OT theory to deal with permutation-invariance in BO allows for effectively and efficiently addressing problems whose decision variables are layouts, namely point-clouds over a continuous space but where the order of points does not affect the objective function. Most of the optimal location problems belong to this setting and, as a proper instance, we have targeted the optimization of the layout of an offshore wind farm.

Empirical results show that permutation-invariance is the key to achieving the best performance against both vanilla BO and baseline methods that do not rely on any probabilistic surrogate model (i.e., different types of sampling and TPE).
Moreover, permutation-invariance also leads to a reduction of the overall computational cost: although a Kantorovitch's problem (linear programming) must be solved at every iteration, the decrease of cost for GP fitting largely overcompensates it.

However, it is important to make the readers aware that in the case of large point-clouds (for instance $m>1000$) the cost for solving the Kantorovich's problem could become problematic. In this case, using some prefixed sorting, instead of OT, can anyway assure permutation-invariance but at the cost of fitting a GP and optimizing the acquisition function on a search space which is no longer box-bounded.

Finally, the proposed approach suggests at least two new research directions: one more methodological, consisting in adopting PIBO for the optimization of the acquisition function in batch-BO, and one more application-driven consisting of extending PIBO for multi-objective optimization, especially for the application considered in this paper.\\

\backmatter
\bmhead{Acknowledgements}
This work is supported by the Eindhoven AI Systems Institute. The authors would also like to thank Aravind Satish from TNO for sharing their knowledge on the practice of offshore wind farm design.\\

\bibliography{mybib}

\newpage 

\begin{appendices}

\section{Description of wind farm ensemble predictor}
\label{sec:surr}

The wind farm ensemble predictor consists of an ensemble of XGBoost models~\cite{chen2016xgboost} that were trained on a dataset~\cite{rawdata} based on the EXPObench library~\cite{EXPObench}.
Here, a wind farm layout optimization problem using the FLORIS simulator~\cite{floris2020} was investigated, with $5$ wind turbines. This simulator takes approximately $15$ seconds per function evaluation, depending on the hardware used.
Five surrogate-based optimization algorithms and random search were applied to this problem, with a budget of $1000$ function evaluations and with $10$ repetitions for each algorithm. Saving all values for the decision variables and objective, for every iteration, gave rise to a dataset of $60.000$ samples, which was already publicly available in~\cite{rawdata}.
The wind farm properties are kept the same as in~\cite{EXPObench}.
The size of the wind farm is $1666.65 \times 1666.65$ m, which was normalized to $[0,1]^2$, and the five wind turbines have a rotor diameter of $126$ m.

After converting the unit of the function evaluations from Wh per year to GWh per year, we sorted all input variables by horizontal position (and by vertical position in case of a tie) to ensure permutation invariance in this ensemble predictor. Then, we first split the dataset into $90\%$ train and $10\%$ test data. After this, we split the train dataset five separate times, with $90\%$ train and $10\%$ validation data. On these five separate splits, we trained five separate XGBoost models, which were identical to each other except for their maximum depth, which was set to $\{54,162,486,1458, 4374\}$ in the different models. Training these models took $4.83, 5.11, 5.22, 5.61$ and $6.48$ seconds respectively on the following hardware: Intel(R) Core(TM) i5-8365U CPU @ 1.60GHz, Ubuntu 24.04.

After training the five models, when predicting a new result, an ensemble was formed that uses the median $y_{med}$ of the five separate predictions. Since the XGBoost model is deterministic, this median value is deterministic too. However, the original FLORIS simulator is stochastic, as it simulates different daily wind conditions to arrive at an average AEP in Wh. To artificially simulate noise in the ensemble, we add random noise to get a stochastic ensemble predictor $\hat y$:
\begin{align}
    \hat y = y_{med} - z |y_{med}-y_{mean}|.
\end{align}

Here, $z\sim U(0,\sigma)$ is a uniform random variable, $\sigma$ is a parameter that we can use to control the noise, which we set to $1$, and $y_{mean}$ is the average of the five predictions. With this noise profile, the variance of the noise becomes larger when the different models in the ensemble disagree on their predictions, leading to a large difference between median and mean. Furthermore, the noise is always negative, which avoids the situation where noise can cause an improvement in the objective value.
With this approach, evaluating the objective function in this ensemble predictor takes only approximately $0.1$ seconds instead of the original $15$ seconds.
We added the ensemble predictor as a problem to the EXPObench library~\cite{EXPObench}.
\end{appendices}

\end{document}